\documentclass{vgtc}                          

\graphicspath{{figures/}{pictures/}{images/}{./}} 

\usepackage{times}                     
\newcommand{\hot}[1]{#1}

\usepackage{amsmath}
\usepackage{tabu}                      
\usepackage{booktabs}                  
\usepackage{lipsum}                    
\usepackage{mwe}                       

\usepackage{mathptmx}                  

\newenvironment{myitemize}{
\begin{itemize}
 \setlength{\itemsep}{1pt}
 \setlength{\parskip}{0pt}
 \setlength{\parsep}{0pt}}{\end{itemize}
}

\onlineid{0}

\vgtccategory{Research}

\title{Exploring \hot{LLM Agent Designs and Interaction Modalities} for\\ Scientific Visualization}

\author{Jackson E. Vonderhorst\thanks{e-mail: jvonder2@nd.edu}\\ %
\scriptsize Univ. Notre Dame
\and Kuangshi Ai\thanks{e-mail: kai@nd.edu}\\ %
\scriptsize Univ. Notre Dame
\and Haichao Miao\thanks{e-mail: miao1@llnl.gov}\\
\scriptsize LLNL
\and Shusen Liu\thanks{e-mail: liu42@llnl.gov}\\
\scriptsize LLNL
\and Chaoli Wang\thanks{e-mail: chaoli.wang@nd.edu}\\
\scriptsize Univ. Notre Dame
}


\begin{document}


\vspace{-0.05in}

\abstract{
This paper examines how large language model (LLM) agents perform on scientific visualization (SciVis) tasks that require generating visualization workflows from natural-language instructions. We compare \hot{three representative agent designs}—domain-specific agents with structured tool use, computer-use agents, and general-purpose coding agents—across 15 benchmark tasks, evaluating visualization quality, efficiency, robustness, computational cost, and the impact of persistent memory. We further study interaction modalities, including code scripts, model context protocol (MCP) or API calls, command-line interfaces (CLI), and graphical user interfaces (GUI). \hot{Our goal is to characterize the tradeoffs among representative SciVis agent configurations used in practice.} The results reveal clear tradeoffs across \hot{agent designs and interaction modalities}. General-purpose coding agents achieve the highest task success rates but incur greater computational cost, whereas domain-specific agents are more efficient and stable but less flexible. Computer-use agents perform well on individual operations but struggle with multi-step workflows. Across both CLI- and GUI-based settings, persistent memory improves performance over repeated trials, but its effectiveness depends on the interaction mode and the quality of feedback. These findings suggest that future SciVis systems should combine structured tool use, interactive capabilities, and adaptive memory mechanisms to balance performance, robustness, and flexibility.
}
\vspace{-0.1in}
\firstsection{Introduction}
\maketitle

\label{sec:intro}

Scientific visualization (SciVis) tools such as ParaView~\cite{Ahrens2005ParaView} are essential for analyzing complex, high-dimensional data, yet they remain difficult to use, particularly for new users. Effective visualization often requires detailed knowledge of rendering pipelines, filter semantics, and scripting APIs, making even simple exploratory tasks time-consuming and error-prone.

Recent advances in large language models (LLMs) have enabled a new generation of autonomous agents capable of translating natural language intent into visualization actions. Systems such as ChatVis~\cite{peterka2025chatvis}, ParaView-MCP~\cite{liu2025paraview}, and AI VIS co-scientist~\cite{miao2026toward} demonstrate that language-driven interfaces can reduce the cognitive burden of interacting with SciVis software. More broadly, multimodal agents now exhibit the ability to perceive visual interfaces, reason about task goals, and execute actions across GUIs designed for human use~\cite{nguyen-etal-2025-gui, zhang2024guiagent}. These developments suggest a shift from manual visualization workflows toward AI-assisted interaction.

Despite this progress, current systems remain unreliable on realistic SciVis tasks, which often require multi-step reasoning, parameter tuning, and perceptual verification. Recent work has called for community collaboration to develop benchmarks that enable meaningful evaluation of agents for scientific data analysis and visualization~\cite{Ai-GenAI25}. SciVisAgentBench~\cite{ai2026svab} addresses this need by providing representative tasks and evaluation protocols, which we adopt to compare \hot{agent configurations} under shared conditions.

%
\hot{
Existing SciVis agents differ along multiple design dimensions, including interaction mechanisms, tool access, planning strategies, memory capabilities, and underlying models. In this work, we examine these systems through the lens of \textit{agent designs} and \textit{interaction modalities}. 
Agent designs describe the capabilities and architectural choices provided by a system. Interaction modalities describe how agents interface with visualization software during task execution.

To facilitate discussion, we group representative systems into three broad categories based on their primary design strategy.}
\textit{Domain-specific agents} (e.g., ChatVis~\cite{peterka2025chatvis}, ParaView-MCP~\cite{liu2025paraview}) operate directly on visualization APIs to achieve efficient and deterministic execution. \textit{Computer-use agents} (e.g., UFO~\cite{zhang-etal-2025-ufo,zhang2025ufo2,zhang2025ufo}, Open Interpreter~\cite{openinterpreter2023}) interact with existing software environments to enable flexible adaptation and error recovery. \textit{General-purpose coding agents} (e.g., Codex CLI~\cite{openai2025codex}, Claude Code~\cite{anthropic2025claudecode}) leverage broad programming capabilities to construct end-to-end workflows with minimal domain constraints. Additionally, some agents (e.g., Agent S~\cite{agashe2024agent}, Letta~\cite{letta2026}) further incorporate persistent memory and reflection to improve performance over repeated trials.

Interaction modality describes how agents execute actions, including \textit{code scripts} and \textit{model context protocol} (MCP) or \textit{API calls} for structured tool use, as well as \textit{command-line interfaces} (CLI) and \textit{graphical user interfaces} (GUI) for more general interaction. 
\hot{These modalities are often associated with particular agent categories but are not themselves controlled experimental factors. More broadly, practical SciVis agents differ simultaneously in interaction mechanisms, tool access, planning strategies, memory capabilities, execution frameworks, and underlying LLM backbones. Consequently, understanding the tradeoffs among various agent configurations remains an important open problem for the SciVis community.}

In this paper, we present a comparative study of \hot{representative agent designs and interaction modalities} for LLM-driven SciVis agents. Rather than proposing a new agent architecture, we focus on evaluating representative systems under realistic tasks derived from SciVisAgentBench. \hot{Our goal is not to isolate the effect of a single design factor, but to understand how representative SciVis agent configurations succeed or fail on realistic tasks and what tradeoffs and design principles may guide future visualization agents.} Specifically, this work makes the following contributions:
\begin{myitemize}
\vspace{-0.05in}
\item \hot{We provide one of the first empirical comparisons of representative SciVis agent configurations 
with an analysis of persistent memory and adaptive learning.}
\item We identify key behavioral tradeoffs, including determinism vs.\ robustness and efficiency vs.\ adaptability, that shape agent effectiveness.
\item We adopt a task-driven evaluation that measures success, consistency, failure modes, and recovery behaviors.
\item We derive design insights suggesting that hybrid agents with deterministic execution, perceptual grounding, and adaptive learning may outperform any \hot{individual agent configuration}.
\vspace{-0.05in}
\end{myitemize}

\vspace{-0.1in}
\section{Related Work}
\label{sec:related}

Recent SciVis research increasingly treats LLMs as agentic systems that execute workflows rather than serve as passive interfaces~\cite{dhanoa2025agentic}. Existing approaches can be organized by \hot{agent designs}, as well as by differences in interaction modality and adaptation capability.

Domain-specific agents translate natural language into structured API calls within visualization systems, e.g., ChatVis~\cite{peterka2025chatvis}, ParaView-MCP~\cite{liu2025paraview}, and InferA~\cite{tam2025infera}, thereby enabling efficient, reproducible execution. However, their tight coupling to predefined tools limits robustness in the face of ambiguous intent or unforeseen states.

Computer-use agents interact directly with software interfaces via perception and low-level actions. 
Domain-specific systems such as AVA~\cite{liu2024ava}, NLI4VolVis~\cite{ai2025nli4volvis}, TexGS-VolVis~\cite{tang2025texgs}, and HiLSVA~\cite{ai2026hilsva} also incorporate visual feedback for iterative refinement, while general frameworks including UFO series~\cite{zhang-etal-2025-ufo, zhang2025ufo2, zhang2025ufo} and Anthropic's computer-use~\cite{anthropic2024computeruse} enable flexible adaptation across applications, which improves error recovery but incurs higher computational cost and variability. Recent work further highlights that effective human oversight in such agents depends on how supervision is structured, rather than simply increasing user control~\cite{chen2026comparing}.
Persistent memory and reflection mechanisms can improve performance over time, as in Agent-S~\cite{agashe2024agent} and OS-Copilot~\cite{wu2024oscopilot}. In SciVis, VizGenie~\cite{biswas2025vizgenie} demonstrates reusable visualization skills. Broader studies emphasize the importance of learning in long-horizon tasks~\cite{fang2025comprehensive, sun2025seagent}, though evaluation remains difficult due to delayed and noisy feedback.

General-purpose coding agents such as Codex~\cite{openai2025codex} and Claude Code~\cite{anthropic2025claudecode} leverage broad programming capabilities to construct workflows end-to-end. While flexible, they often incur higher costs and lack the structured guarantees of domain-specific systems. \hot{Recent studies further show that domain-specific agent skills can improve coding-agent performance on SciVis tasks by providing reusable procedural knowledge and tool-specific guidance~\cite{ai2026scivisagentskills}.}

Evaluating agent behavior remains challenging. Prior work considers execution success, perceptual similarity, and user studies. CoDA~\cite{chen2025coda}, MatPlotBench~\cite{yang2024matplotagent}, and VisEval~\cite{chen2024viseval} propose complementary metrics, while benchmarks such as VisualWebArena~\cite{koh2024visualwebarena} and WindowsAgentArena~\cite{bonatti2024windows} extend evaluation to broader environments. Beyond outcome evaluation, recent HCI systems emphasize real-time monitoring and analysis of intermediate states. Tools such as VizCode~\cite{yang2024vizcode} and SPARK~\cite{yang2025spark} enable fine-grained tracking of multi-step programming processes, underscoring the importance of inspecting intermediate states and workflow structure rather than relying solely on final outputs. In SciVis, recent efforts call for more systematic evaluation frameworks~\cite{Ai-GenAI25}. SciVisAgentBench~\cite{ai2026svab} provides realistic tasks and outcome-based protocols, while SVLAT~\cite{do2026svlat} introduces a standardized assessment for scientific visualization literacy. However, existing benchmarks do not explicitly analyze how \hot{agent designs and interaction modalities} influence robustness, consistency, and failure modes in \hot{complex multi-step} workflows.

\vspace{-0.1in}
\section{Experimental Setup}
\label{sec:experiment}

\hot{We evaluate representative SciVis agent systems through a controlled comparison of different agent designs and interaction strategies operating on shared ParaView workflows.}
Rather than proposing a new architecture, we compare three \hot{representative agent configurations}: {\em domain-specific agents}, {\em computer-use agents}, and {\em general-purpose coding agents}, and additionally analyze the effect of persistent memory in selected agents. We examine differences in effectiveness, robustness, computational cost, and behavioral patterns.

Our study evaluates agents on tasks derived from SciVisAgentBench~\cite{ai2026svab}, which provides realistic visualization workflows representative of common ParaView usage. All agents are tested under identical task specifications and system environments to enable direct comparison. 
Our experiments focus not only on whether agents complete tasks, but also on how they behave during execution, including their ability to recover from errors, adapt to unexpected states, and maintain consistency across repeated trials.

{\bf Agent settings.}\
We evaluate eight representative agents spanning \hot{different agent design strategies and interaction modalities}. 
Domain-specific agents include ChatVis~\cite{peterka2025chatvis} and ParaView-MCP~\cite{liu2025paraview}, which operate through scripts or structured API calls. 
Computer-use agents include the Microsoft UFO series~\cite{zhang-etal-2025-ufo, zhang2025ufo2, zhang2025ufo} and Open Interpreter~\cite{openinterpreter2023}, which interact with software through GUIs.
General-purpose coding agents include Claude Code~\cite{anthropic2025claudecode} and Codex~\cite{openai2025codex}, which primarily operate via CLI-based code generation.
In addition, Agent~S~\cite{agashe2024agent} and Letta~\cite{letta2026} are evaluated in both learning-enabled and learning-disabled configurations to isolate the effect of persistent memory across GUI- and CLI-based settings.
\hot{For Agent-S, memory consists of previously observed task trajectories and execution experiences that can be retrieved to guide future actions. For Letta, memory consists of persistent records of prior interactions, execution outcomes, and planning context that can be recalled during subsequent tasks.}

\begin{figure}[htb]
\centering
\includegraphics[width=\linewidth]{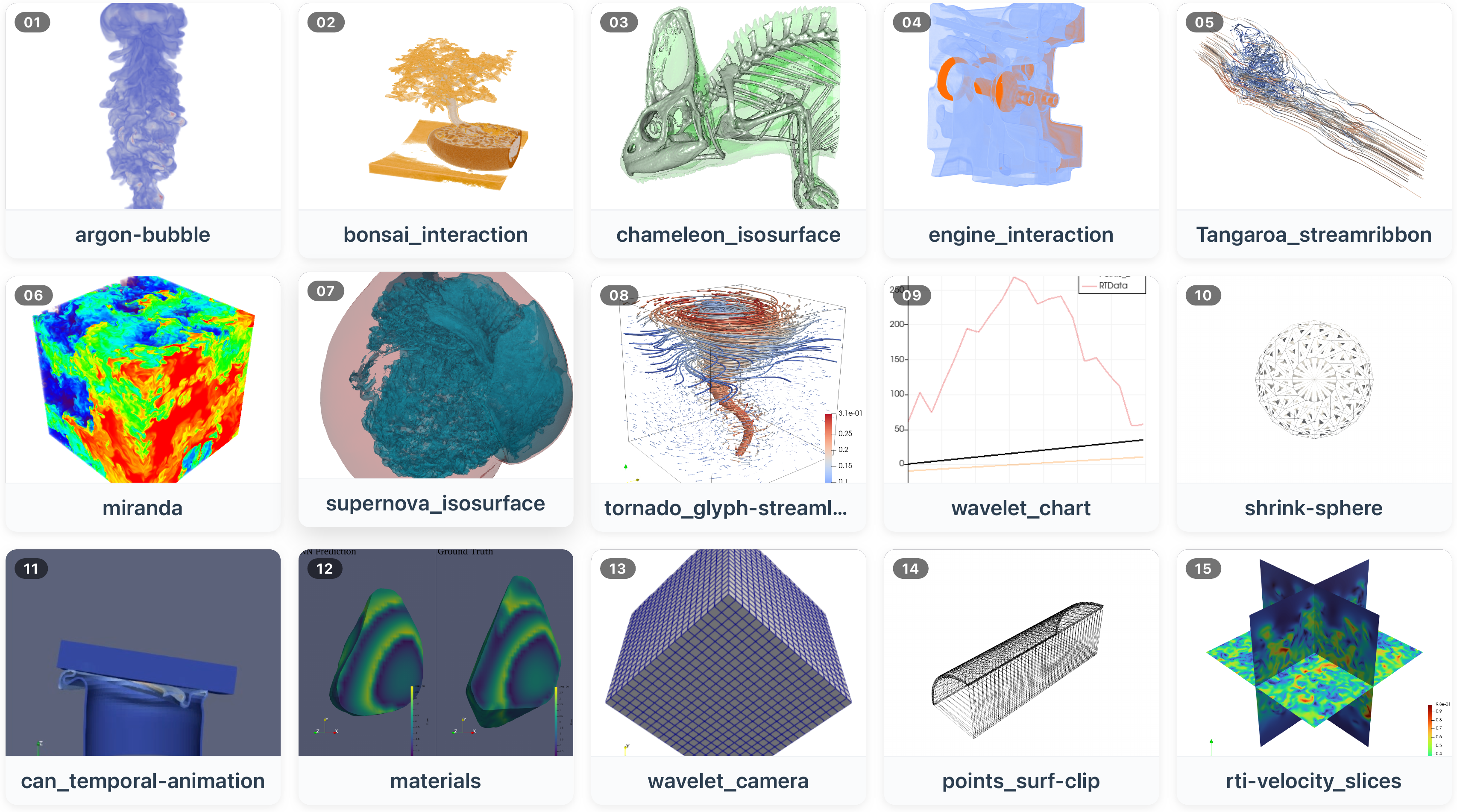}
\vspace{-0.25in}
\caption{The 15 representative ParaView visualization tasks from SciVisAgentBench.}
\label{fig:example_cases}
\end{figure}

{\bf Task design.}\
All agents are evaluated on a shared subset of 15 ParaView tasks derived from SciVisAgentBench (Figure~\ref{fig:example_cases}), \hot{selected for a representative coverage of the SciVisAgentBench taxonomy across application domains, data types, visualization operations, and complexity levels~\cite{ai2026svab}, while keeping the experimental cost manageable.} 
\hot{These tasks represent common ParaView workflows and span multiple levels of complexity, ranging from single-step operations (e.g., applying a filter or loading a dataset) to multi-step visualization pipelines involving dependent operations such as filter composition, parameter tuning, camera configuration, and output generation.}
Each of the eight evaluated agents is tested across the full set of tasks, with each task executed 10 times per agent to capture performance variability and assess consistency across repeated trials.

{\bf Stepwise evaluation.}\
For agents operating through GUIs (UFO, Open Interpreter, and Agent-S), we additionally evaluate step-by-step task decomposition. Each task is manually broken into a sequence of intermediate steps, and a corresponding gold-standard state is defined for each step. Agents execute each step independently, and outputs are compared against the gold-standard state using the same rubric criteria as in full-task evaluation. A step is marked as a pass if the generated state sufficiently matches the expected intermediate result, and a failure otherwise. To control for compounding errors, each step begins from the gold-standard state of the previous step rather than the agent's prior output. This isolates the agent's ability to perform individual operations correctly. Final performance is reported as the proportion of correctly executed steps across all tasks and trials.

\begin{table*}[t]
\centering
\caption{Benchmark performance and resource usage on the 15 representative ParaView visualization tasks. The overall score is reported on a 0-100 scale as mean$\pm$std across 10 trials. Token counts and time are per-task statistics, averaged over all 15 tasks across 10 trials (mean$\pm$std). \hot{Completion rate measures the proportion of runs that finish without explicit execution errors. In contrast, pass metrics are based on the LLM-judge (Claude-Opus-4.6) rubric, in which a trial passes if its score is at least 50\%.} See Figure~\ref{fig:passk_curves} for pass@k and pass$^{\wedge}$k curves. 
}
\label{tab:overall_results}
\vspace{-0.1in}
\resizebox{1.0\textwidth}{!}{%
\begin{tabular}{lcccccl}
\toprule
Setting & Overall Score $\uparrow$ & Completion Rate $\uparrow$ & Input Tokens $\downarrow$ & Output Tokens $\downarrow$ & Time $\downarrow$ & \hot{Agent Configuration} \\
\midrule
ChatVis+GPT-5.2 & 37.25$\pm$4.92 & 45.0\%$\pm$8.5\% & 8.17K$\pm$842 & 10.81K$\pm$816 & 1m 35s$\pm$16s & Domain-Specific (Code Script) \\
ParaView-MCP+Claude Sonnet-4.6 & 30.49$\pm$7.08 & 50.7\%$\pm$8.9\% & 146.67K$\pm$17.89K & 7.82K$\pm$739 & 2m 11s$\pm$15s & Domain-Specific (MCP) \\
Codex CLI+GPT-5.2 & 68.99$\pm$1.92 & 100.0\%$\pm$0.0\% & 774.52K$\pm$623.29K & 7.48K$\pm$4.54K & 2m 15s$\pm$1m 27s & General-Purpose Coding (CLI) \\
Claude Code+Claude-Sonnet-4.6 & 66.35$\pm$3.92 & 96.0\%$\pm$4.7\% & 346.44K$\pm$340.56K & 7.78K$\pm$7.44K & 2m 49s$\pm$2m 47s & General-Purpose Coding (CLI) \\
UFO+GPT-5.2 & 15.71$\pm$4.52 & 27.1\%$\pm$10.7\% & 310.96K$\pm$29.46K & 5.84K$\pm$422 & 6m 13s$\pm$1m 3s & Computer-Use (GUI) \\
Open Interpreter+Claude-Sonnet-4.6 & 14.04$\pm$6.03 & 27.9\%$\pm$11.2\% & 211.37K$\pm$15.14K & 4.26K$\pm$443 & 5m 31s$\pm$59s & Computer-Use (GUI) \\
Agent-S (Learning Enabled)+GPT-5.2 & 18.31$\pm$8.97 & 40.0\%$\pm$12.6\% & 242.00K$\pm$18.53K & 3.74K$\pm$318 & 5m 22s$\pm$52s & Memory+Learning (GUI) \\
Agent-S (Learning Disabled)+GPT-5.2 & 10.75$\pm$6.11 & 28.6\%$\pm$9.8\% & 276.90K$\pm$20.10K & 5.07K$\pm$480 & 5m 58s$\pm$1m 8s & Memory+Learning (GUI) \\
Letta (Learning Enabled)+Claude-Sonnet-4.6 & 30.78$\pm$9.44 & 42.9\%$\pm$10.3\% & 11.51K$\pm$801 & 8.77K$\pm$841 & 1m 47s$\pm$10s & Memory+Learning (CLI) \\
Letta (Learning Disabled)+Claude-Sonnet-4.6 & 19.09$\pm$6.42 & 32.9\%$\pm$6.8\% & 18.17K$\pm$1.88K & 10.24K$\pm$914 & 2m 35s$\pm$26s & Memory+Learning (CLI) \\
\bottomrule
\end{tabular}}
\vspace{-0.2in}
\end{table*}

\begin{figure}[t]
\centering
$\begin{array}{c@{\hspace{0.01in}}c}
\includegraphics[width=0.475\linewidth]{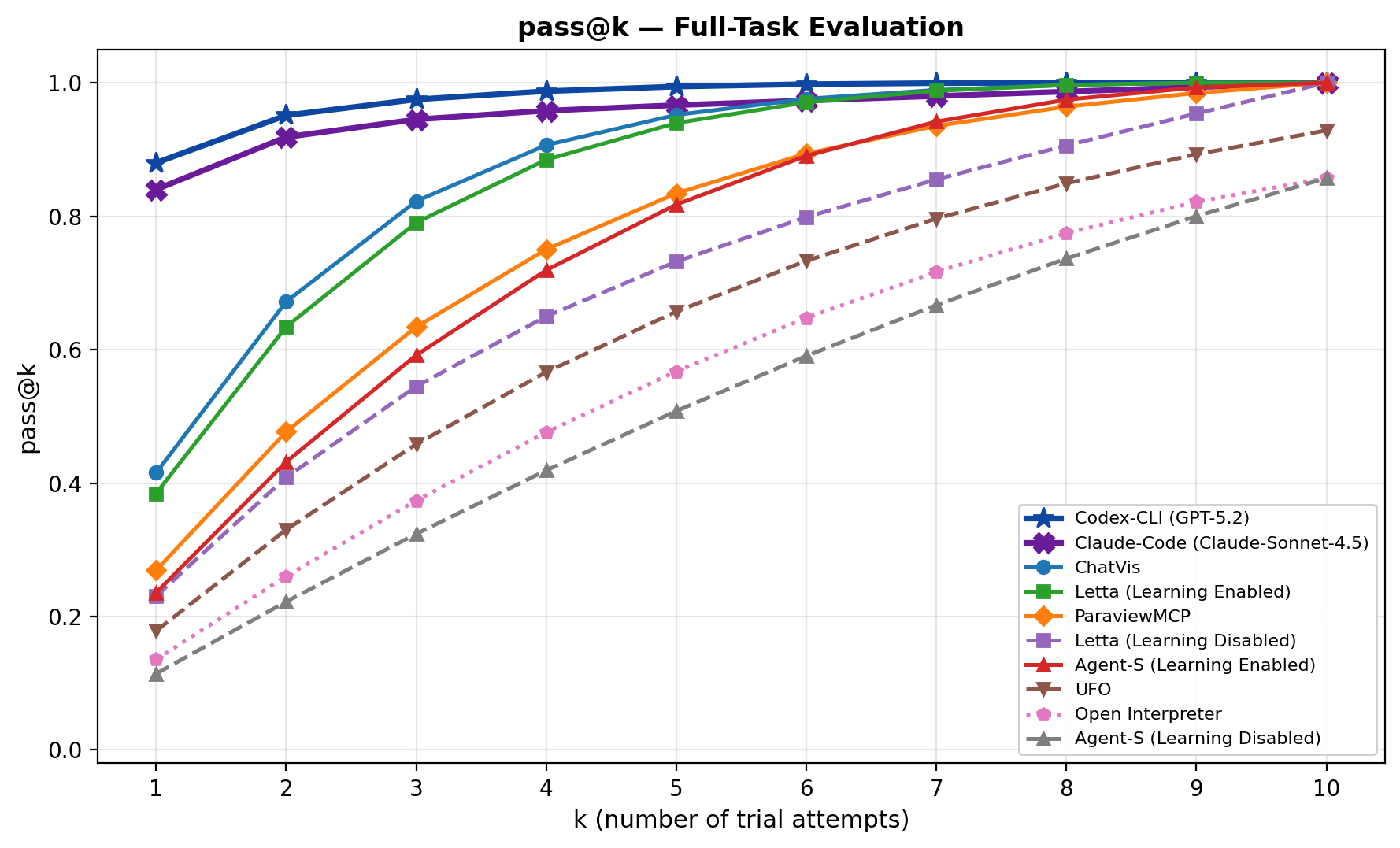}&
\includegraphics[width=0.475\linewidth]{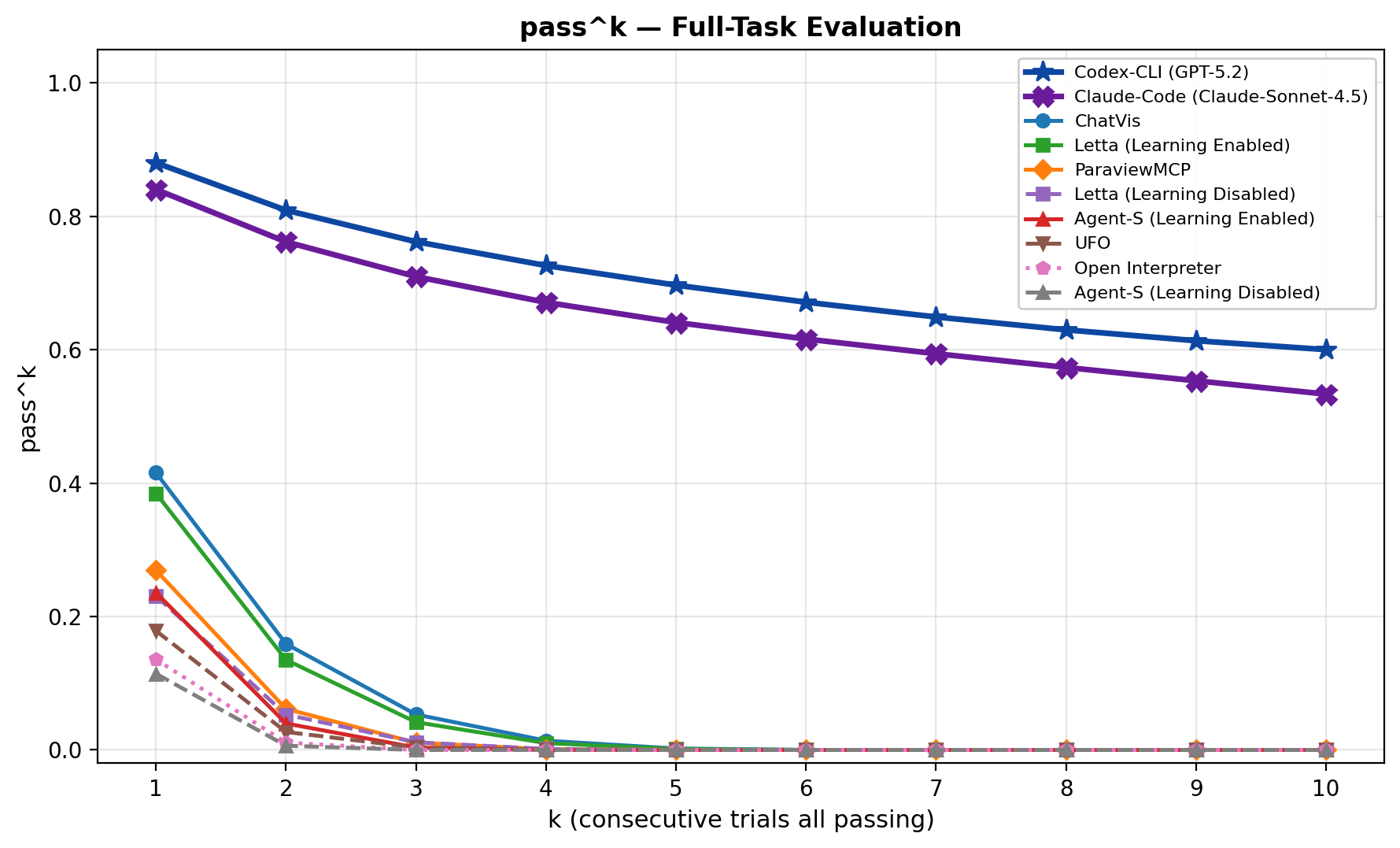}\\
\end{array}$
\vspace{-0.15in}
\caption{pass@k (top) and pass$^{\wedge}$k (bottom) curves for full-task evaluation. ChatVis and Letta (Learning Enabled) lead, with all coding agents converging to pass@k $\approx$ 1.0 by k=10. pass$^{\wedge}$k decays sharply, with only the top performers retaining non-zero values past k=4. 
}
\label{fig:passk_curves}
\end{figure}

{\bf Analysis goals.}\
Our goal is not to identify a universally superior agent. Domain-specific agents are expected to excel in efficiency, while computer-use agents may offer greater adaptability. Persistent memory and adaptive learning may further improve some agents with repeated experience. Instead, we aim to understand which capabilities are most beneficial for SciVis and how different \hot{agent configurations} succeed or struggle in realistic workflows.
Specifically, this study investigates: (1) the reliability of each \hot{agent configuration} across repeated trials; (2) the behavioral factors that distinguish successful from unsuccessful executions; (3) the efficiency tradeoffs including runtime and token usage; (4) the impact of persistent memory and learning on \hot{multi-step} task performance; and (5) the extent to which complementary strengths may inform the design of future SciVis agents.

\begin{table}[t]
\centering
\caption{Stepwise evaluation of GUI-based agents. Each of the 15 tasks is decomposed into 5 steps, yielding $15 \times 5$ step instances per trial. At each step, the agent starts from the ground-truth state of the previous step and executes the next. We run 3 trials. Step Pass Rate is computed per trial as $\#\text{successful steps}/(15 \times 5)$ and averaged across trials (mean$\pm$std). Token counts and time are per-task statistics, averaged over 15 tasks across 3 trials (mean$\pm$std). 
}
\label{tab:stepwise_results}
\vspace{-0.1in}
\resizebox{\columnwidth}{!}{%
\begin{tabular}{lcccc}
\toprule
Setting & Input Tokens $\downarrow$ & Output Tokens $\downarrow$ & Time $\downarrow$ & Step Pass Rate $\uparrow$ \\
\midrule
UFO+GPT-5.2 & 296.93K$\pm$50.69K & 7.36K$\pm$596 & 5m 44s$\pm$37s & 63.9\%$\pm$5.9\%  \\
Open Interpreter+Claude-Sonnet-4.6 & 199.46K$\pm$43.90K & 5.42K$\pm$420 & 5m 12s$\pm$32s & 60.7\%$\pm$6.7\%  \\
Agent-S (Learning Enabled)+GPT-5.2           & 221.54K$\pm$38.23K & 5.38K$\pm$461 & 5m 3s$\pm$31s  & 65.2\%$\pm$8.9\% \\
Agent-S (Learning Disabled)+GPT-5.2           & 221.54K$\pm$38.23K & 5.38K$\pm$461 & 5m 3s$\pm$31s  & 59.5\%$\pm$8.1\% \\
\bottomrule
\end{tabular}%
}
\end{table}

\vspace{-0.1in}
\section{Results and Discussion}

Consistent patterns emerge across all experiments. General-purpose coding agents achieve the highest task completion rates and pass@k performance, but at significantly higher token usage and runtime compared to all other \hot{configurations}. Domain-specific agents exhibit substantially lower computational cost and more stable execution patterns, though they remain limited in flexibility and overall performance ceiling. Computer-use agents underperform on full-task workflows but achieve substantially higher success rates under stepwise decomposition, indicating that multi-step planning, rather than interface perception, is their primary limitation. Stepwise decomposition improves both execution quality and computational efficiency by constraining planning complexity and reducing token usage and runtime. Finally, in auxiliary comparisons on Agent-S and Letta, persistent memory and learning improve both performance and efficiency by reducing redundant exploration of unsuccessful strategies.

\vspace{-0.075in}
\subsection{Overall Performance and Cost-Accuracy Tradeoffs} 

We evaluate all eight agents on 15 SciVisAgentBench tasks, with results summarized in Table~\ref{tab:overall_results} and pass@k / pass$^{\wedge}$k curves shown in Figure~\ref{fig:passk_curves}. General-purpose coding agents operating via CLI achieve the strongest overall performance. Codex CLI attains the highest overall score (68.99$\pm$1.92) and a 100.0\% completion rate, followed closely by Claude Code (66.35$\pm$3.92, 96.0\% completion). Both models also dominate pass@k, approaching saturation by $k=10$.
However, this performance comes at a substantial computational cost. Codex CLI requires an average of 774.52K input tokens per task, with high variance, while Claude Code similarly incurs large token overhead. These results indicate that CLI-based agents achieve high reliability through extensive code generation and iterative refinement, rather than solely through efficient task execution.
In contrast, domain-specific agents (ChatVis and ParaView-MCP) exhibit substantially lower cost profiles, with input token usage reduced by one to two orders of magnitude (e.g., 8.17K for ChatVis). Although their completion rates (45.0\%--50.7\%) remain below those of general-purpose coding agents, they demonstrate greater efficiency per successful execution, reflecting the advantages of structured tool access and constrained action spaces.
Taken together, these results establish a clear tradeoff: general-purpose coding agents maximize task completion, whereas domain-specific agents optimize efficiency.

\vspace{-0.075in}
\subsection{Consistency and Pass@k Behavior} 

The pass@k and pass$^{\wedge}$k curves in Figure~\ref{fig:passk_curves} provide additional insight into execution reliability. While several agents achieve high pass@k values, particularly at larger $k$, pass$^{\wedge}$k declines rapidly across all models, indicating that consistent success across repeated trials remains limited.
Even the strongest general-purpose coding agents exhibit variability in execution trajectories, suggesting that their success often depends on favorable intermediate decisions rather than fully stable reasoning strategies. Domain-specific agents and Letta (learning enabled) demonstrate stronger early pass@k performance, reflecting more deterministic execution patterns. However, their asymptotic performance remains lower, highlighting a tradeoff between consistency and ultimate task-completion capability.

\vspace{-0.075in}
\subsection{Limitations of GUI-Based Interaction} 

Computer-use agents (UFO, Open Interpreter, and Agent-S) underperform in the full-task setting, with completion rates below 40\%, substantially higher runtime (5--6 minutes per task), and token usage (200K--300K range) relative to other \hot{configurations} (Table~\ref{tab:overall_results}).
This performance gap is largely attributable to the demands of GUI-based interaction. Unlike script- or API-driven agents, these models must repeatedly interpret high-dimensional visual input, maintain \hot{execution context across multiple steps}, and execute fine-grained interface actions. These requirements introduce compounding sources of error, particularly in multi-step workflows where early inaccuracies propagate through later actions.
%

However, these results should not be interpreted as a fundamental limitation of GUI-based agents. SciVis pipelines often require iterative visual verification, parameter-sensitive operations, and maintenance of intermediate visualization states. Errors in transfer functions, camera placement, pipeline ordering, or rendering parameters may remain visually ambiguous until later stages of the workflow, causing failures to propagate across operations. As a result, \hot{multi-step SciVis workflows} place disproportionate demands on perceptual grounding and sequential reasoning, both of which remain challenging for current multimodal agents.

\vspace{-0.075in}
\subsection{Stepwise Decomposition and Interaction Granularity} 

To isolate the effect of \hot{multi-step workflow execution}, we evaluate GUI-based agents under stepwise task decomposition (Table~\ref{tab:stepwise_results}). Under this setting, step-level pass rates increase to approximately 60\%--65\% across all models.
In addition to improving completion rates, stepwise decomposition also yields consistent reductions in token usage and runtime. Because each step constrains the action space and reduces planning depth, agents avoid redundant exploration and repeated correction cycles that are common in full-task execution. This suggests that a significant portion of the inefficiency observed in GUI-based agents arises from the overhead of \hot{multi-step} reasoning rather than the cost of individual actions.
GUI-based agents are capable of accurate localized reasoning, but struggle to maintain coherence across extended action sequences. In other words, the primary limitation of computer-use agents is not interface understanding itself, but planning depth. When task complexity is reduced through decomposition, their performance becomes substantially more competitive at the step level.
GUI-based agents may be better suited for interactive or human-in-the-loop settings, where task structure can be incrementally specified, rather than for fully autonomous execution of long SciVis workflows.

More broadly, these results suggest that future SciVis agents may benefit from hierarchical workflow decomposition and intermediate-state supervision. Visualization workflows naturally expose semantically meaningful intermediate states that can be inspected, validated, and corrected before execution continues. This creates opportunities for mixed-initiative interaction and stepwise evaluation strategies that are particularly well suited to SciVis environments.

\vspace{-0.0775in}
\subsection{Effects of Persistent Memory and Adaptive Learning} 

We further evaluate the impact of persistent memory by comparing learning-enabled and learning-disabled configurations for Letta and Agent-S (Table~\ref{tab:overall_results}).
\hot{The evaluated memory mechanisms primarily store and retrieve prior execution experiences, enabling agents to reuse successful strategies and avoid repeating previously unsuccessful behaviors.}
Across both Letta and Agent-S, enabling memory yields consistent improvements in completion rate, overall score, and efficiency. For example, Letta improves from 19.09 to 30.78 in overall score, while Agent-S improves from 10.75 to 18.31.
Notably, the benefits of persistent memory extend beyond improvements in success rate. Across both Letta and Agent-S, enabling memory also reduces token usage and execution time. This indicates that persistent memory enables agents to avoid revisiting previously unsuccessful strategies, leading to more efficient trajectories through the solution space. In this sense, memory improves effectiveness and efficiency without introducing additional computational overhead.
These gains indicate that persistent memory reduces the occurrence of repeated failures and enables more efficient exploration of the solution space. In the case of Letta, the improvements are particularly pronounced, as memory reduces redundant script generation and repeated execution errors. For Agent-S, the gains are more modest, since perceptual and interaction overhead remain dominant even when experience is retained.
At the same time, the coding-based memory setting reveals an important limitation for SciVis tasks. An agent may converge to workflows that are syntactically valid and executable, yet semantically incorrect from a visualization perspective. Correctness in SciVis often depends on perceptual properties of rendered outputs, such as spatial structure, feature visibility, and camera configuration. Persistent memory alone is insufficient and should be complemented by visual validation and SciVis-specific feedback mechanisms.

\vspace{-0.075in}
\subsection{\hot{Synthesis, Implications, and Limitations}}

Taken together, these results suggest that future SciVis agents should not be designed as purely autonomous code generators or GUI operators. Effective SciVis systems may instead require support for visualization-state reasoning, perceptual verification, and hierarchical workflow decomposition. SciVis workflows frequently contain semantically incorrect yet executable intermediate states, making visual feedback and intermediate validation particularly important. Our findings further suggest that evaluation protocols for SciVis agents should extend beyond final task completion to include intermediate-state correctness, perceptual consistency, and robustness across \hot{complex multi-step} workflows. More broadly, the results indicate that no single \hot{configurations} is sufficient, while hybrid systems combining structured tool use, perceptual grounding, and adaptive memory mechanisms may provide the most promising direction for future SciVis agents. \hot{Meanwhile, the evaluated systems differ along multiple dimensions, including backbone models, planning strategies, memory mechanisms, tool access, and interaction interfaces. Consequently, the observed performance differences should be interpreted as tradeoffs among representative SciVis agent configurations rather than the causal effect of any single design factor. Future work could conduct controlled ablation studies to isolate the contribution of individual components.}

\vspace{-0.1in}
\section{Conclusions and Future Work}
\label{sec:conclusions}

We present a comparative study of \hot{agent designs and interaction modalities} for LLM-based SciVis agents, evaluating domain-specific, computer-use, and general-purpose coding agents on realistic multi-step SciVis tasks, together with an auxiliary analysis of persistent memory in selected agents. Our results show that while general-purpose coding agents achieve the highest task completion rates, they incur significantly higher computational costs. Domain-specific agents provide more efficient and stable execution but lack flexibility. In contrast, computer-use agents reveal strong step-level reasoning capabilities \hot{but struggle to maintain coherent execution across multi-step workflows}. Persistent memory improves both effectiveness and efficiency by reducing redundant exploration across repeated trials.



Looking forward, two primary directions emerge for future SciVis agent design. First, continued improvements in general-purpose coding agents suggest that CLI-based frameworks may become increasingly powerful for fully automated workflows. \hot{Their key advantage lies in the flexibility of operating within a full programming environment rather than through predefined tool interfaces. As foundation models become more capable, the reliability gap between coding agents and structured tool-use agents may continue to narrow.} At the same time, introducing domain-specific constraints through procedural guidelines, such as agent skills~\cite{anthropic2025agentskillsblog}, could substantially improve efficiency and reliability by reducing redundant exploration. Second, hybrid approaches that combine complementary \hot{agent configurations} offer a promising solution. For example, future agents may integrate deterministic API-based execution for reliable low-level operations, CLI-based reasoning for complex workflow generation, GUI interaction for perceptual grounding, and persistent memory mechanisms to adapt execution strategies over time.

A key challenge for future work is enabling reliable SciVis-specific self-evaluation mechanisms. Unlike conventional programming tasks, visualization correctness often depends on perceptual and semantic properties that cannot be fully verified through code execution alone. In particular, integrating visual feedback into learning loops could allow agents to assess intermediate visualization states and detect issues such as incorrect transfer functions, camera configurations, and feature visibility, improving both accuracy and consistency. Additionally, incorporating selective elements of GUI-based interaction may provide useful grounding for debugging and recovery in complex workflows.
More broadly, future research should explore how to balance efficiency, robustness, and flexibility across \hot{agent designs and interaction modalities}, and how to design unified systems that combine structured tool use, perceptual grounding, and adaptive memory mechanisms for real-world scientific workflows. 

\vspace{-0.1in}
\acknowledgments{
This research was supported by the U.S.\ NSF through grants IIS-2101696, OAC-2104158, and IIS-2401144, and the U.S.\ DOE through grants DE-SC0023145 and ECRP 51917/SCW1885.
This work was performed under the auspices of the DOE by Lawrence Livermore National Laboratory under contract DE-AC52-07NA27344, reviewed and released under LLNL-CONF-2019246. 
The authors thank the anonymous reviewers for their insightful comments.}

\vspace{-0.1in}
\bibliographystyle{abbrv-doi-hyperref-narrow}

\bibliography{refs-abbv}
\end{document}